\documentclass{article}

\PassOptionsToPackage{round,authoryear}{natbib}

\usepackage[preprint]{neurips_2026}

\usepackage{textcomp}
\usepackage{amsmath,amssymb}
\usepackage{multirow}
\usepackage[font=small,labelfont=bf,skip=6pt]{caption}
\usepackage[section,above]{placeins}   
\usepackage{adjustbox}
\usepackage{graphicx}
\usepackage{xurl}
\usepackage{tikz}
\usetikzlibrary{positioning,calc,arrows.meta,shapes.geometric}
\usepackage{fvextra}   
\fvset{fontsize=\scriptsize, frame=single, framesep=2mm, rulecolor=\color{gray!60}, breaklines=true, breaksymbolleft={}}
\usepackage{enumitem}
\usepackage{booktabs}

\usepackage[capitalise,noabbrev]{cleveref}

\newcommand{\jevact}{JevAct}
\newcommand{\passhat}[1]{pass\textasciicircum{}#1}
\newcommand{\passat}[1]{pass@#1}

\title{Think Before You Act:\\ Decision Gate for LLM Agents}

\author{
Iftach Shoham\thanks{Equal contribution.} \\
IBM SIL
\And
Yuval David\footnotemark[1] \\
IBM SIL
\And
Offer Akrabi \\
IBM SIL
\And
Erez Bilgory \\
IBM SIL
\And
Ido Levy \\
IBM SIL
\And
Sami Marreed \\
IBM SIL
\And
Sergey Zeltyn \\
IBM SIL
\And
Segev Shlomov\thanks{Corresponding author: segev.shlomov1@ibm.com.} \\
IBM SIL
}

\begin{document}
\maketitle

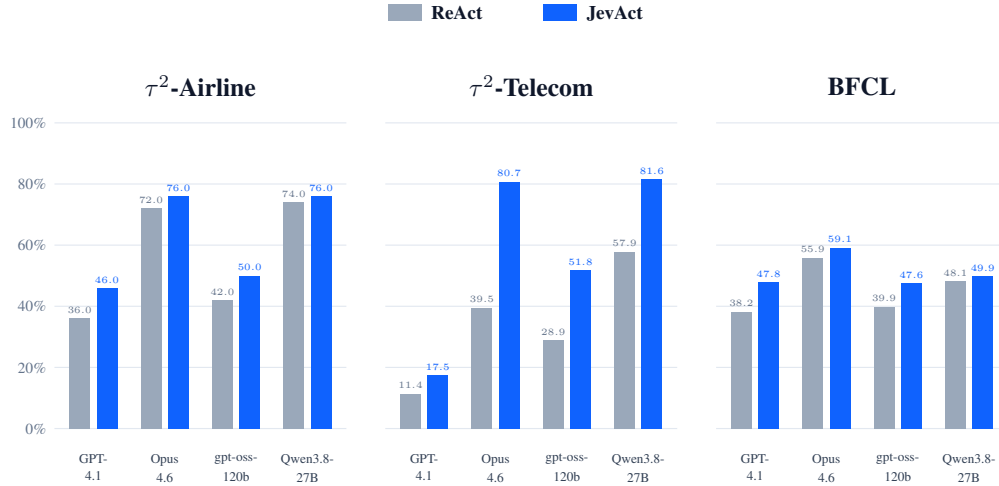
\begin{figure}[!ht]
\centering
\makebox[\textwidth][c]{\resizebox{\textwidth}{!}{\definecolor{passreact}{HTML}{9AA8BA}
\definecolor{passjevact}{HTML}{0F62FE}
\definecolor{passink}{HTML}{0F172A}
\definecolor{passmuted}{HTML}{334155}
\definecolor{passgrid}{HTML}{E2E8F0}
\definecolor{passaxis}{HTML}{64748B}
\definecolor{passreactvalue}{HTML}{64748B}
\definecolor{passjevactvalue}{HTML}{0F62FE}

\begin{tikzpicture}[x=0.035in,y=0.0165in,font=\sffamily]
\tikzset{every node/.style={inner sep=0pt}}

\newcommand{\scorescale}{0.75}

\newcommand{\scorevalue}[1]{%
  \pgfmathprintnumber[
    fixed,
    fixed zerofill,
    precision=1
  ]{#1}%
}


\newcommand{\scorebar}[4]{%
  \fill[#3]
    ({#1-1.6},0) rectangle ({#1+1.6},#2);
  \node[
    anchor=south,
    text=#4,
    font=\fontsize{5}{6}\selectfont,
    align=center
  ] at (#1,{#2+1.6})
  {\scalebox{\scorescale}{\scorevalue{#2}}};
}

\newcommand{\jevbar}[2]{%
  \fill[passjevact]
    ({#1-1.6},0) rectangle ({#1+1.6},#2);
  \node[
    anchor=south,
    text=passjevactvalue,
    font=\bfseries\fontsize{5}{6}\selectfont,
    align=center
  ] at (#1,{#2+1.6})
  {\scalebox{\scorescale}{\scorevalue{#2}}};
}

\newcommand{\modelpair}[4]{%
  \node[
    anchor=north,
    align=center,
    text=passmuted,
    font=\fontsize{5.2}{5.8}\selectfont
  ] at (#1,-8) {#2};

  \scorebar{#1-2.1}{#3}{passreact}{passreactvalue}
  \jevbar{#1+2.1}{#4}
}

\node[
  fill=passreact,
  minimum width=0.18in,
  minimum height=0.09in
] at (54,136) {};

\node[
  anchor=west,
  text=passink,
  font=\bfseries\fontsize{7.5}{8}\selectfont
] at (58,136) {ReAct};

\node[
  fill=passjevact,
  minimum width=0.18in,
  minimum height=0.09in
] at (78,136) {};

\node[
  anchor=west,
  text=passink,
  font=\bfseries\fontsize{7.5}{8}\selectfont
] at (82,136) {JevAct};

\newcommand{\panelgrid}[2]{%
  \foreach \v in {0,20,40,60,80,100}{
    \draw[passgrid,line width=0.45pt]
      (#1,\v) -- (#2,\v);
  }
}

\node[
  text=passink,
  font=\bfseries\fontsize{10}{11}\selectfont
] at (22.5,112) {$\tau^2$-Airline};

\panelgrid{0}{45}

\foreach \v in {0,20,40,60,80,100}{
  \node[
    anchor=east,
    text=passaxis,
    font=\fontsize{6}{7}\selectfont
  ] at (-1.2,\v) {\v\%};
}

\modelpair{6}{\shortstack{GPT-\\4.1}}{36.0}{46.0}
\modelpair{17}{\shortstack{Opus\\4.6}}{72.0}{76.0}
\modelpair{28}{\shortstack{gpt-oss-\\120b}}{42.0}{50.0}
\modelpair{39}{\shortstack{Qwen3.8-\\27B}}{74.0}{76.0}

\node[
  text=passink,
  font=\bfseries\fontsize{10}{11}\selectfont
] at (73.5,112) {$\tau^2$-Telecom};

\panelgrid{51}{96}

\modelpair{57}{\shortstack{GPT-\\4.1}}{11.4}{17.5}
\modelpair{68}{\shortstack{Opus\\4.6}}{39.5}{80.7}
\modelpair{79}{\shortstack{gpt-oss-\\120b}}{28.9}{51.8}
\modelpair{90}{\shortstack{Qwen3.8-\\27B}}{57.9}{81.6}

\node[
  text=passink,
  font=\bfseries\fontsize{10}{11}\selectfont
] at (124.5,112) {BFCL};

\panelgrid{102}{147}

\modelpair{108}{\shortstack{GPT-\\4.1}}{38.2}{47.8}
\modelpair{119}{\shortstack{Opus\\4.6}}{55.9}{59.1}
\modelpair{130}{\shortstack{gpt-oss-\\120b}}{39.9}{47.6}
\modelpair{141}{\shortstack{Qwen3.8-\\27B}}{48.1}{49.9}

\end{tikzpicture}}}
\caption{\passhat{3} results from Table~\ref{tab:main}.}
\label{fig:main-pass3-shared}
\end{figure}

\begin{abstract}
LLM agents increasingly perform consequential actions through external tools, yet their proposed actions are often executed without independent verification of policy compliance or task completion. We introduce \textbf{JevAct}, a runtime verification framework that intercepts selected agent actions before execution and enables feedback-driven correction. JevAct compiles domain policies into executable state checks and structured questions evaluated by a prefill-only decision model, which produces probability-based judgments without relying on LLM generation. By combining deterministic verification of tool-state facts with probabilistic assessment of conversational context, the framework identifies potentially invalid actions and premature task termination, providing targeted feedback for bounded revision. JevAct operates independently of the underlying agent model and requires no modifications to its parameters or tools. We evaluate JevAct across eleven LLM agents on $\tau^2$-bench Airline and Telecom tasks and BFCL. Results show improvements in agent reliability across multiple models and benchmarks, with particularly consistent gains on Telecom and BFCL, while performance on Airline is more mixed. Overall, these findings suggest that combining structured decision models with executable verification can improve the reliability of tool-using LLM agents.
\end{abstract}

\section{Introduction}
\label{sec:intro}

Language-model agents increasingly use tools to change records, make bookings, and guide users through multi-step procedures. A mistaken answer can sometimes be corrected in a later turn; an unauthorized write or a premature hand-off may be harder to undo. Yet many agent loops execute a proposed tool call as soon as the model produces it. This leaves a gap between generating an action and checking that it is appropriate to carry out.

ReAct agents interleave reasoning with tool use, while reflection methods such as Self-Refine and Reflexion add critique and feedback to improve later proposals or trials~\citep{yao2022react,madaan2023self,shinn2023reflexion}. These methods can improve agent behavior, but they do not by themselves verify a particular action against a domain policy immediately before execution. Runtime constraints and learned tool-call checks address this separate problem~\citep{wang2026agentspec,chen2025shieldagent,mou2026toolsafe,kang2026policyguard}.

A separate generative LLM could review each proposal and explain its concerns, as in LLM-as-a-judge methods~\citep{zheng2023judging,liu2023g}. But free-form explanations must still be interpreted before a gate can act on them, and the cost of making this decision with a frontier model is high. Decision models instead return choices and probabilities for fixed questions, making threshold-based checks straightforward. Their probabilities are not guaranteed to be calibrated or correct, so their value has to be tested in the agent loop rather than assumed~\citep{typesafe2026jev,deusser2026evaluating}.

We present \jevact{}, a harness that checks selected proposals before execution. Offline, a compiler turns a domain policy into questions, rules, executable checks over tool state, and where needed, a hand-off checklist. At runtime, code checks facts available in tool records and Kev (an open sourced Jev-like~\citep{typesafe2026jev} decision model) answers questions that depend on the conversation. A passing proposal proceeds; a flagged proposal is returned to the agent with targeted feedback and a bounded chance to revise. The agent still proposes actions; Kev does not choose them. This separates \jevact{} from decision-centric runtimes that delegate action selection to a decision model~\citep{jevloop2026}.

Our central question in this is work is whether a decision model can serve as a probability-thresholded verifier alongside executable tool-state checks. We compare \jevact{} with a ReAct harness using the same agent LLM in both arms, across Airline and Telecom tasks in $\tau^2$-bench and multi-turn function-calling tasks in BFCL~\citep{barres2025tau,patil2025berkeley}. On Telecom, all eight agents improve both at-least-once success and success across all three runs. On BFCL, all nine improve in three-run success, but at-least-once results are mixed. Airline effects are smaller and vary by agent (see Figure~\ref{fig:main-pass3-shared}). Because \jevact{} adds compiled checks and revision opportunities, these comparisons evaluate the full harness, not the judge in isolation.

Our contributions in this work are threefold:
\begin{itemize}
   \item We introduce \jevact{}, which combines executable checks over tool state with thresholded decision-model judgments of conversational context before selected actions are executed.
   
    \item We lay the foundation for a workflow that compiles domain policies into questions, decision rules, executable state checks, and hand-off checklists for a decision-model gate.
    
    \item We evaluate the harness against same-model ReAct baselines across three benchmark settings, with component and judge ablations alongside task-success, repeatability, and estimated judge-cost results.
\end{itemize}

\section{Related Work}
\label{sec:related}

\subsection{LLM Agents and Test-Time Feedback}
ReAct interleaves model reasoning with environment actions~\citep{yao2022react}. Self-Refine and Reflexion add test-time critique or feedback to improve later proposals and trials without updating model weights~\citep{madaan2023self,shinn2023reflexion}. These methods motivate giving an agent a chance to revise, but do not by themselves check whether a consequential tool call satisfies a domain policy before execution.

\subsection{Runtime Policy Enforcement}
Several systems place explicit controls around agent execution. GuardAgent uses a separate agent to generate and execute verification code from user-defined guard requirements~\citep{xiang2024guardagent}. Zwerdling et al. compile company policies into executable guards that check compliance before tool actions~\citep{zwerdling2025towards}. AgentSpec represents runtime constraints as triggers, predicates, and enforcement actions; ShieldAgent derives verifiable rules from policy documents; and CUGA applies governance interventions across an execution graph~\citep{wang2026agentspec,chen2025shieldagent,shlomov2026governance}. ToolSafe uses a learned guardrail to assess proposed tool calls and return feedback, while PolicyGuard and PolicyGuide use model-based verifiers for policy adherence and workflow guidance~\citep{mou2026toolsafe,kang2026policyguard,kang2026policyguide}. Related approaches include verifier-grounded runtime mediation, which uses executable procedural checks to select interventions~\citep{yabusaki2026verifier}, and ToolFence, which combines deterministic authorization with judge-based capability grants~\citep{li2026toolfence}. CaMeL instead enforces security through separation of trusted control flow and untrusted data~\citep{debenedetti2026defeating}. JevLoop assigns action and control decisions to a decision model; \jevact{} leaves proposals to the ReAct LLM and places Kev, an open-source Jev-type decision model, in a separate verification step before selected actions commit~\citep{jevloop2026}. We focus on probability-thresholded verification using a fixed-choice decision model, combined with executable checks of tool-state facts and feedback-driven revision.

\subsection{LLM Judges and Decision Models}
LLM-as-a-judge methods use generative models to compare responses, as in MT-Bench, or apply explicit rubrics, as in G-Eval; such judgments can also be sensitive to response order~\citep{zheng2023judging,liu2023g,wang2024large}. Jev introduced decision models that return typed choices, scores, and probabilities rather than generated text~\citep{typesafe2026jev}. Initial evaluations study its closed-label decisions and rubric-based scoring~\citep{deusser2026evaluating}. Two recent preprints evaluate Jev as a judge: Li et al. report competitive results on preference and evidence-grounded factuality but weaker performance when derivation is required, while Rao and Callison-Burch report mixed rubric results and overlapping errors with LLM judges~\citep{li2026jev,rao2026jev}. Those studies judge completed outputs. \jevact{} instead uses Kev, an open-source Jev-type model, online before selected actions execute, combining its fixed-choice judgments with programmatic state checks and feedback for revision~\citep{kev}.

\section{Method}
\label{sec:method}

In this work we introduce \jevact{}, which augments the harness of a ReAct agent~\citep{yao2022react} by inserting a selective checking stage between message proposal and execution. The checks can defer execution and request a revised proposal
from the LLM with temporary feedback, subject to a fixed revision budget.
The agent's LLM, base prompt, and tools remain unchanged.

\subsection{Overview}
\label{sec:overview}

\paragraph{Setting.}
At each step $t$, an LLM reads the conversation history and proposes a message, which may contain user-facing text, tool
calls, or both. The harness normally executes the proposal and adds it and any tool results to the history. In
policy-governed tasks, \jevact{} checks selected proposals before execution because some mistakes are hard to undo. A \emph{write call} changes the environment, for example by booking, cancelling, or changing a setting. A
\emph{hand-off} ends the agent's attempt, either by transferring the user to a human or by telling the user that
nothing more can be done. The same runtime design can also apply generic checks when no written policy is available.

\begin{figure}[t]
\centering
\begin{adjustbox}{max width=\textwidth}
\begin{tikzpicture}[
  font=\footnotesize,
  node distance=6mm and 10mm,
  box/.style={draw, rounded corners=2pt, align=center, inner sep=4pt, minimum height=11mm, fill=white},
  off/.style={box, fill=gray!10},
  chk/.style={box, fill=blue!7, align=left},
  dec/.style={draw, diamond, aspect=2.2, inner sep=1pt, align=center, fill=white},
  arr/.style={-{Stealth[length=2mm]}, semithick},
  lab/.style={font=\scriptsize, inner sep=1.5pt},
  num/.style={circle, draw, fill=white, inner sep=0.6pt, minimum size=3.6mm, font=\scriptsize\bfseries},
  note/.style={font=\scriptsize\itshape, text=gray},
]
\node[box] (hist) {History\\$h_t$};
\node[box, right=of hist] (llm) {\textbf{LLM} $\pi$\\proposes $m$};
\node[dec, right=of llm] (trig) {write call\\or hand-off?};
\node[chk, right=of trig] (check) {\textbf{Check}\\[1pt]
  code: list items, compute facts\\
  judge (Kev): yes/no per item\\
  rule: pass or flag};
\node[box, right=of check] (commit) {\textbf{Commit}\\execute $m$};

\draw[arr] (hist) -- (llm);
\draw[arr] (llm) -- (trig);
\draw[arr] (trig) -- node[lab, above]{yes} (check);
\draw[arr] (check) -- node[lab, above]{pass} (commit);
\draw[arr] (trig.south) -- ++(0,-6mm) -| node[lab, pos=0.25, below]{no: as in ReAct} (commit.south);
\draw[arr] ($(check.north west)!0.4!(check.north east)$) -- ++(0,6mm) -|
  node[lab, pos=0.25, above]{flag: discard $m$, return feedback $f$ (shown once)} (llm.north);

\node[num] at (llm.north west) {1};
\node[num] at (trig.north) {2};
\node[num] at (check.north west) {3};
\node[num] at (commit.north west) {4};

\node[off, above=21mm of check] (compiled) {\textbf{Checks}\\questions, rules, code,\\hand-off checklist};
\node[off, left=8mm of compiled] (compiler) {\textbf{Compiler}\\(LLM)};
\node[off, left=8mm of compiler] (policy) {Policy $\Pi$, tool schemas,\\unlabeled logs};
\draw[arr] (policy) -- (compiler);
\draw[arr] (compiler) -- (compiled);
\draw[arr, dashed] (compiled) -- (check);

\node[note, anchor=south west] at (policy.north west) {offline};
\node[note, anchor=north west] at ([yshift=-9mm]hist.south west) {online, at every step of the agent};
\end{tikzpicture}
\end{adjustbox}
\caption{\jevact{}. Offline (top), an LLM compiles the policy into checks. Online (bottom), the agent's proposal goes
through four stages: \textbf{(1)}~propose, \textbf{(2)}~trigger, \textbf{(3)}~check when required, and \textbf{(4)}~commit or
revise. Proposals without a write call or hand-off are executed as in ReAct. A flagged proposal is discarded, and the
LLM proposes again with temporary feedback.}
\label{fig:overview}
\end{figure}
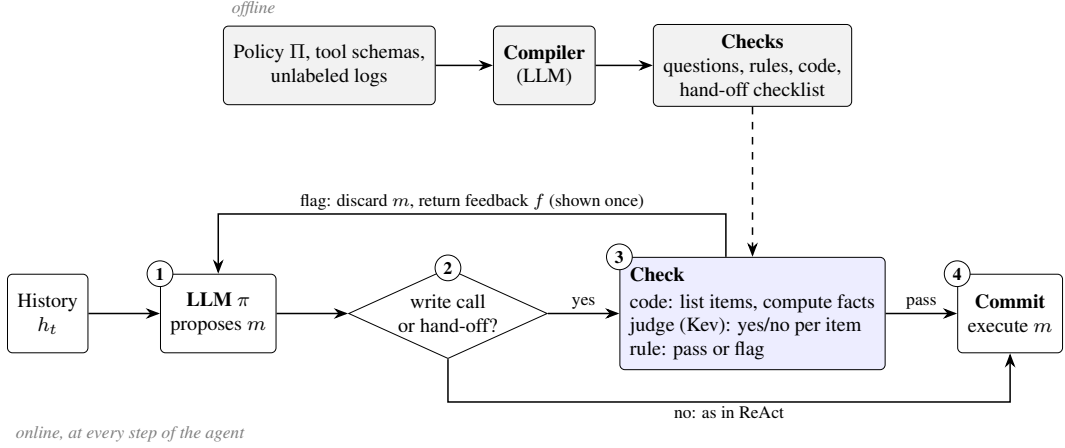

\paragraph{Components.}
Our method \jevact{} has four main components, as illustrated in Figure~\ref{fig:overview}.
(1)~\textbf{Decision loop}: Similar to a ReAct agent, every step is composed of four stages: \emph{propose},
\emph{trigger}, \emph{check}, and \emph{commit or revise}. The trigger selects a check for a write call or hand-off;
other proposals proceed as in ReAct. A passing proposal is committed, while a flagged one is returned to the agent
with feedback so it can revise.
(2)~\textbf{Compiler}: Offline, an LLM turns each domain's policy into yes/no questions, rules that combine their
answers, code that computes facts from tool results, and (where needed) a hand-off checklist.
(3)~\textbf{Judge}: A decision model answers the questions with probabilities, allowing a check to act only on
confident answers.
(4)~\textbf{Checks}: The \emph{write gate} tests whether a proposed write call is allowed by the policy. The
\emph{hand-off check} looks for required troubleshooting steps that remain undone. For write calls, code computes
what can be established from tool records; the judge answers questions that depend on the conversation. The hand-off
check uses the judge to assess which problem applies and which checklist steps are settled. Although the components rely on each other, each is modular and can be independently replaced or improved in future work.

\subsection{Decision loop}
\label{sec:loop}

At step $t$, the agent goes through four stages:
\begin{enumerate}[leftmargin=*,itemsep=2pt]
  \item \textbf{Propose.} The LLM proposes a message $m \sim \pi(\cdot \mid h_t)$ given the history $h_t$.
  \item \textbf{Trigger.} If $m$ contains a write call, the write gate applies. If $m$ hands off, the hand-off check
        applies. Any other message is committed directly.
  \item \textbf{Check.} The check returns \emph{pass} or \emph{flag}. A flag comes with a feedback text $f$ that names
        the problem.
  \item \textbf{Commit or revise.} A passed message is executed and added to the history, as in ReAct. A flagged
        message is discarded: it is never executed and never stored. The LLM then proposes again with the feedback
        appended for this one call, $m' \sim \pi(\cdot \mid h_t, f)$.
\end{enumerate}
Each check has a revision budget: $R = 2$ for the write gate and $R = 1$ for the hand-off check. While budget remains,
a flagged proposal is discarded and the agent gets another chance to respond to the feedback. Once the budget is
exhausted, the next proposal is committed without another check. Thus, a check cannot block the agent indefinitely,
and it adds at most $R$ extra LLM calls. Feedback is temporary and is not added to the conversation history, so
critiques of earlier steps do not accumulate in the context.

\subsection{Compiler}
\label{sec:compile}

The compiler runs offline for each domain. It uses the policy and tool schemas to build the checks; code generation
and repair also use example tool outputs and calls from development logs. For each write tool, it produces:
\begin{itemize}[leftmargin=*,itemsep=2pt]
  \item yes/no questions about policy requirements, the source of call arguments, user intent, and any required prior
        steps or confirmations;
  \item a Boolean rule that combines the answers and specifies when the call is allowed; and
  \item a Python function that computes policy-relevant facts available in tool records (such as dates, amounts, and
        counts) and returns any violations those records establish.
\end{itemize}
For hand-offs, it also produces the problem types covered by the policy and the steps the agent should try for each
problem before giving up. Each step has a \emph{settled} question: was it tried, verified, or ruled out?
One instruction in the compile prompt matters most: \emph{a condition that was never discussed is not ruled out}.
Without it, ``enable roaming'' compiles to ``\ldots{} or the user is not traveling'', which looks settled even though
nobody asked. With it, the checklist gets its own step: ask whether the user is traveling.

For write gates, the compiler gets one repair pass using logged calls without outcome labels. Next, questions that the
judge answers with a confident violation on more than 10\% of a tool's development calls are dropped; this filters
questions that are hard to answer from the available evidence. The hand-off checklist has a separate filter: steps
flagged on more than 5\% of hand-offs in solved training episodes are dropped. The prompts and examples are in Section~\ref{app:compile}.

\subsection{Judge}
\label{sec:judge}

For the judge in this work, we use Kev~\citep{kev}, an open-source Jev-like decision model. Given a prompt and questions with fixed answer options, it returns
a probability for each option in one forward pass, without generating text. Questions about the same prompt share
that pass. We use Kev-27B, a fine-tune of Qwen3.8-27B, without further training for our setup. Its probabilities let the checks act
only on confident answers, and the cost of a check does not grow with the number of questions. Because Kev is separate
from the agent LLM, the agent does not judge its own proposals. The checks divide the work by evidence source: code
calculates facts available in tool records, while Kev answers focused questions that require interpreting the
conversation, such as whether the user requested an action or whether a step was tried. This asks the judge to assess
specific candidate items rather than diagnose an entire problem from scratch. Other models that answer the same questions can take Kev's place.
We chose Kev as open source model.

\subsection{Checks}
\label{sec:checks}

\paragraph{Write gate.}
The write gate runs before each write call, in three steps:
\begin{enumerate}[leftmargin=*,itemsep=2pt]
  \item \textbf{Code:} the compiled function reads the proposed call's arguments and earlier tool results, then returns
        any violations established by those records.
  \item \textbf{Judge:} Kev answers the questions for that tool using the policy, conversation, tool results (with
        dates annotated and list values ranked), and proposed call. An answer counts as \emph{no} only when Kev is
        confident, meaning $P(\text{yes}) \le 0.05$; otherwise it counts as \emph{yes}.
  \item \textbf{Rule:} the call is flagged if the code finds a violation or the Boolean rule is not satisfied by the
        question answers.
\end{enumerate}
Feedback lists the established violations and failed questions, then asks the agent to correct the call, ask the user
for missing information, or explain why the request cannot be fulfilled. If the agent believes the call is already
correct, it may submit it again. After a second flag, the feedback directs the agent to respond to the user rather
than call the tool again.

\paragraph{Hand-off check.}
The hand-off check runs before a transfer call and before a message that gives up in words. A keyword filter first
identifies possible give-up messages, which the judge then confirms. From the conversation, Kev identifies the user's
problem, whether the user says it is fixed, and whether each checklist step is settled. A step is \emph{open} when
the probability that the user's problem requires it is at least $0.5$ and the probability that it is settled is at
most $0.5$. If the problem is not fixed and at least one required step is open, the check flags the hand-off and
returns the open steps as feedback. The hand-off check allows one revision; after that budget is used, the next
proposal is committed without another check.

\paragraph{Function calling without a policy.}
BFCL multi-turn~\citep{patil2025berkeley} has no written policy, and its user turns are fixed, so we use two generic checks.
The \emph{end-of-turn check} runs when the agent ends its turn while a tool error stating a requirement (for example,
``needs to'' or ``must'') remains unresolved. Kev selects the tool that meets the requirement from the tool list, or
selects none. If the probability of a tool being the right one is at least $0.6$, feedback directs the agent to call
it rather than ask the user. The \emph{requested-action gate} runs before each state-changing call and asks whether the
user requested it or a tool error requires it. If Kev assigns at most $0.05$ probability to ``yes,'' the gate flags
the call as unrequested and directs the agent to do only what was asked or required.

\section{Experimental Setup}
\label{sec:setup}

\subsection{Benchmarks}
\label{sec:benchmarks}
\textbf{$\tau^2$-bench}~\citep{barres2025tau}: an agent serves a simulated user under a domain policy
and calls tools that read and write a database. We use two domains. \emph{Airline} (50 tasks) tests policy-compliant
write calls (book, cancel, change). \emph{Telecom} (the 114-task base split) tests troubleshooting, in which the agent
also guides the user through actions on their own device; here the main failure is giving up too early. The user
simulator is GPT-4.1, as in $\tau^2$-bench. Scoring is $\tau^2$-bench's own: the final database state and the required
communication.

\textbf{BFCL multi-turn}~\citep{patil2025berkeley}: 800 entries, 200 in each of four categories (base, missing function,
missing parameter, long context). The user turns are fixed, so there is no user simulator. We follow BFCL's
function-calling loop (no system prompt, up to 20 steps per user turn) and score with its own multi-turn checker: an
entry is solved only if every turn's end state and response match the ground truth. BFCL has no written policy, so
\jevact{} uses its two generic checks here, the end-of-turn check and the requested-action gate (Section~\ref{sec:checks}).

\subsection{Agent LLMs}
\label{sec:llms}
We evaluate eleven agent LLMs. GPT-4.1, Claude Opus~4.6, gpt-oss-20b, gpt-oss-120b, Qwen3.8-27B, and DeepSeek-V4-Pro
are tested on all three benchmark settings. Claude Haiku~4.5 and Claude Opus~5.5 are tested on Telecom only; Llama~3.3
70B, Mistral Small~3.2~24B, and Gemma~4~31B are tested on BFCL only (Table~\ref{tab:main}). Within each comparison, the
ReAct and \jevact{} arms use the same agent model. Agent calls use temperature 0 and a maximum of 4,096 output tokens;
the GPT-4.1 user simulator on $\tau^2$-bench also runs at temperature 0.

\subsection{Judge and Compiler}
\label{sec:setup-judge}
Unless stated otherwise, every \jevact{} result uses the same configuration. The judge is Kev-27B, used as is, for
every agent LLM and benchmark. Claude Opus 4.6 compiles the checks offline for policy-governed tasks (Section~\ref{sec:compile}).
The write gate counts an answer as \emph{no} only when $P(\text{yes}) \le 0.05$ and allows $R = 2$ revisions. The
hand-off check is active only on Telecom, with $R = 1$. On BFCL, the end-of-turn check acts at $P \ge 0.6$ and the
requested-action gate at $P \le 0.05$. For the main results and judge-size ablation, thresholds and dropped items were
set with Kev-27B. In the judge-choice ablation, each alternative LLM judge uses its own label-free dropped-question
list. Table~\ref{tab:kevsize} varies judge size with the checks and thresholds fixed; Table~\ref{tab:judge-choice} also changes the dropped-question list for alternative LLM judges.

\subsection{Metrics}
\label{sec:metrics}
Main results and judge-comparison runs use three runs per task. The single-trial Airline component ablation is reported as \passhat{1} (Table~\ref{tab:ablation}). With $c$ of $n$
runs of a task solved, we report, averaged over tasks~\citep{chen2021evaluating,yao2024tau},
\[
  \text{\passat{k}} = 1 - \frac{\binom{n-c}{k}}{\binom{n}{k}}, \qquad \text{\passhat{k}} = \frac{\binom{c}{k}}{\binom{n}{k}}.
\]
With $n = k = 3$, \passat{3} is the share of tasks solved in at least one run, and \passhat{3} is the share solved in all
three runs. \passhat{1} is the mean success rate across individual runs. Cost is reported as tokens per episode for
the agent LLM and judge, along with the number of proposals sent back (Appendix able~\ref{tab:tokens}).

\subsection{Checks Per Benchmark}
\Cref{tab:instantiation} summarizes which checks are active on each benchmark. The hand-off check is disabled on
$\tau^2$-bench Airline because, in offline tests, it flagged 26\% of hand-offs in otherwise solved episodes.

\begin{table}[t]
\centering
\small
\caption{Checks per benchmark, with the number of compiled items (dropped items in parentheses).}
\label{tab:instantiation}
\begin{tabular}{@{}lll@{}}
\toprule
Benchmark & Check & Compiled items \\
\midrule
$\tau^2$ Airline & write gate & 6 write tools, 86 questions (17) \\
$\tau^2$ Telecom & write gate & 6 write tools, 75 questions (5) \\
                 & hand-off check & 8 problem types, checklist steps (7 dropped) \\
BFCL             & end-of-turn check & -- \\
                 & requested-action gate & -- \\
\bottomrule
\end{tabular}
\end{table}

\section{Results}
\label{sec:results}

\Cref{tab:main} compares \jevact{} with ReAct using the same agent LLM in each pair. The results vary by benchmark: the strongest and most consistent gains appear on Telecom, while Airline and BFCL show a more mixed pattern.

On $\tau^2$-Telecom, all eight agents improve on both metrics. Averaged across agents, pass@3 rises by 16.0 percentage points and \passhat{3} by 16.9 points. The largest change is for Opus~4.6, whose success across all three runs rises from 39.5\% to 80.7\%. GPT-4.1's pass@3 rises from 45.6\% to 83.3\%. Even the strongest ReAct baseline, Opus~5.5, improves in \passhat{3} from 74.6\% to 82.5\%. The Telecom hand-off check is designed to catch early abandonment, so this pattern is consistent with that check addressing a frequent failure mode. The comparison evaluates the full harness, however, and does not isolate the hand-off check's effect.

On $\tau^2$-Airline, gains are smaller and mixed. Pass@3 improves for four of six agents, is unchanged for one, and falls for one. Success across all three runs improves for four agents and falls for two. GPT-4.1's pass@3 rises from 66.0\% to 82.0\%, while Opus~4.6 rises from 84.0\% to 92.0\%. With 50 tasks, a single task changes a score by 2 percentage points, so small differences should be read cautiously. The write gate also intervenes infrequently for some agents: it sends back 0.08 drafts per run for Opus~4.6 and 0.03 for Qwen3.8-27B (Table~\ref{tab:tokens}).

On BFCL, \passhat{3} increases for all nine agents, by 0.1 to 9.6 points. Pass@3 improves for four agents and declines for five, by 0.2 to 3.6 points. Thus, higher success across all three runs does not mean that the gate expands task coverage for every agent; the pattern is instead consistent with more repeatable performance on some tasks.

\begin{table}[tp]
\centering
\small
\caption{Main results (\%). \passat{3}: tasks solved in at least one of three runs. \passhat{3}: tasks solved in all three
runs. Both arms use the same LLM and differ only in the harness; the higher value of each pair is in \textbf{bold}.}
\label{tab:main}

\begin{tabular}{@{}ll cc cc@{}}
\toprule
 & & \multicolumn{2}{c}{\passat{3}} & \multicolumn{2}{c}{\passhat{3}} \\
\cmidrule(lr){3-4} \cmidrule(l){5-6}
Benchmark & Model & ReAct & \jevact{} & ReAct & \jevact{} \\
\midrule
\multirow{6}{*}{\begin{tabular}[c]{@{}l@{}}$\tau^2$ Airline \\\end{tabular}} & GPT-4.1 & 66.0 & \textbf{82.0} & 36.0 & \textbf{46.0} \\
 & Opus 4.6 & 84.0 & \textbf{92.0} & 72.0 & \textbf{76.0} \\
 & gpt-oss-20b & 66.0 & \textbf{70.0} & \textbf{48.0} & 44.0 \\
 & gpt-oss-120b & \textbf{68.0} & 66.0 & 42.0 & \textbf{50.0} \\
 & Qwen3.8-27B & 90.0 & 90.0 & 74.0 & \textbf{76.0} \\
 & DeepSeek-V4-Pro & 86.0 & \textbf{90.0} & \textbf{68.0} & 66.0 \\
\midrule
\multirow{8}{*}{\begin{tabular}[c]{@{}l@{}}$\tau^2$ Telecom \\ \end{tabular}} & GPT-4.1 & 45.6 & \textbf{83.3} & 11.4 & \textbf{17.5} \\
 & Haiku 4.5 & 47.4 & \textbf{69.3} & 19.3 & \textbf{31.6} \\
 & Opus 4.6 & 72.8 & \textbf{100.0} & 39.5 & \textbf{80.7} \\
 & Opus 5.5 & 99.1 & \textbf{100.0} & 74.6 & \textbf{82.5} \\
 & gpt-oss-20b & 83.3 & \textbf{92.1} & 28.1 & \textbf{34.2} \\
 & gpt-oss-120b & 78.9 & \textbf{93.9} & 28.9 & \textbf{51.8} \\
 & Qwen3.8-27B & 91.2 & \textbf{98.2} & 57.9 & \textbf{81.6} \\
 & DeepSeek-V4-Pro & 86.8 & \textbf{96.5} & 53.5 & \textbf{68.4} \\
\midrule
\multirow{9}{*}{\begin{tabular}[c]{@{}l@{}}BFCL \\ \end{tabular}} & GPT-4.1 & 54.0 & \textbf{60.2} & 38.2 & \textbf{47.8} \\
 & Opus 4.6 & 63.2 & \textbf{66.5} & 55.9 & \textbf{59.1} \\
 & gpt-oss-20b & \textbf{54.2} & 53.6 & 33.6 & \textbf{33.8} \\
 & gpt-oss-120b & \textbf{57.8} & 54.2 & 39.9 & \textbf{47.6} \\
 & Qwen3.8-27B & \textbf{65.9} & 64.9 & 48.1 & \textbf{49.9} \\
 & DeepSeek-V4-Pro & \textbf{57.1} & 54.9 & 41.4 & \textbf{45.1} \\
 & Llama 3.3 70B & \textbf{12.2} & 12.0 & 10.5 & \textbf{11.5} \\
 & Mistral Small 3.2 24B & 26.0 & \textbf{27.9} & 15.5 & \textbf{17.1} \\
 & Gemma 4 31B & 62.5 & \textbf{63.2} & 61.5 & \textbf{61.6} \\
\bottomrule
\end{tabular}

\end{table}

\FloatBarrier

\subsection{Ablations}
\label{sec:ablations}

We examine three aspects of \jevact{}: the contribution of its verification components, the trade-off between decision models and generative LLM judges, and the effect of judge size.

\paragraph{Write-gate components.}
Table~\ref{tab:ablation} examines the contribution of different write-gate components using a single-trial Airline evaluation with GPT-4.1 as the agent. The full gate improves success from 50.0\% to 72.0\%, while executable checks alone achieve 68.0\%. In contrast, Kev-27B questions alone reach 57.1\%, suggesting that programmatic verification accounts for much of the improvement in this setting. Replacing Kev-27B with the smaller Kev-9B reduces success to 58.0\%, indicating that judge choice can substantially affect performance. Reducing the revision budget from two to one has a relatively small effect (68.0\%). Importantly, flagging every write call once without checking its validity gives no improvement over ReAct (50.0\%), with three runs per task and a gpt-oss-120b user simulator, an unconditional re-check of every write leaves GPT-4.1's \passhat{3} unchanged (26.0\% to 28.0\%) while the full gate raises it to 40.0\% (paired 95\% CI of the gain over ReAct: $+2.0$ to $+26.0$ points); for gpt-oss-120b as the agent, the same re-check, or random flags at the gate's rate, raise \passhat{3} as much as the full gate (32.0\% to 42.0\%), so whether the judge's verdicts matter beyond the revision opportunity depends on the agent. The flag rates further show that the full gate intervenes on fewer than half of write calls, while achieving the highest success rate among the tested configurations. As these results are based on a single trial, differences between configurations should be interpreted cautiously.

\suppressfloats[t]
\begin{table}[tp]
\centering
\small
\caption{Single-trial write-gate ablation on Airline with GPT-4.1 as the agent. The experiment uses the same configuration as Table~\ref{tab:main} (code checks, Kev-27B, and $R=2$).}
\label{tab:ablation}
\begin{tabular}{@{}l c c@{}}
\toprule
 & \passhat{1} & Flags per task \\
\midrule
ReAct & 50.0 & -- \\
\jevact{} (same configuration as Table~\ref{tab:main}) & 72.0 & 0.38 \\
Code checks only & 68.0 & 0.28 \\
Kev-27B questions only & 57.1 & 0.29 \\
Kev-9B instead of Kev-27B & 58.0 & 0.22 \\
One revision instead of two & 68.0 & 0.32 \\
Placebo: every write flagged once & 50.0 & 1.00 \\
\bottomrule
\end{tabular}

\end{table}

\paragraph{Judge choice and cost.}
\Cref{tab:judge-choice} compares Kev with open-weight and API-based generative LLM judges. All judged configurations improve \passhat{1} over their corresponding ReAct baselines by 9.4--18.7 percentage points, suggesting that the verification framework is not dependent on a particular judge architecture. Kev-27B achieves competitive performance: with GPT-4.1 as the agent, it reaches 70.0\% success, compared with 67.3--72.0\% for generative judges; with Haiku~4.5, it reaches 68.0\%, compared with 64.7--66.0\%. The main advantage of Kev is efficiency. Its estimated judge cost ranges from \$0.0003 to \$0.0017 per task, compared with \$0.0020--\$0.0040 for open-weight LLM judges and \$0.0127--\$0.0231 for API judges. Thus, prefill-only decision models can provide competitive runtime verification at substantially lower estimated judge cost, without requiring autoregressive generation. These estimates exclude the agent's own inference cost; Section~\ref{app:judge-cost} provides further details.

\paragraph{Judge size.}
Table~\ref{tab:kevsize} evaluates whether increasing the decision model's size improves verification. Surprisingly, larger judges do not consistently yield better downstream task performance. On Airline, Kev-27B performs best for GPT-4.1 (65.3\% \passhat{1}, compared with 59.3\% for Kev-9B and 58.0\% for Kev-0.8B), but the remaining agents show no consistent preference for larger models. On Telecom, smaller judges frequently match or outperform Kev-27B, indicating that increased model capacity does not necessarily translate into more successful agent interactions. Moreover, in a replay of 609 identical hand-off states, Kev-9B agrees with Kev-27B on 97.0\% of decisions, while Kev-0.8B agrees on 90.6\% (Section~\ref{app:offline-judges}). This suggests that substantially smaller decision models can often reproduce the decisions of a larger judge.

One factor affecting judge size is input length. On Airline, 41\% of write-check inputs exceed 8{,}192 tokens, beyond the smaller judges' validated context length. Offline analysis shows that Kev-9B misses more invalid calls on long inputs, while threshold recalibration reduces the gap on shorter inputs. Kev-0.8B remains weaker even on short inputs (Section~\ref{app:judge-inputs}, and~\ref{app:write-judges}). Together, these findings suggest that effective judge selection depends not only on model capacity, but also on context length, decision thresholds, and the type of verification required. Since downstream success also depends on the agent's response to feedback, these results should not be interpreted as direct rankings of judge accuracy.

\paragraph{Overall findings.}
The ablations highlight three observations. First, executable state checks provide a substantial part of the measured benefit on Airline, and selective verification beats indiscriminate revision for GPT-4.1 but not for gpt-oss-120b, for which a re-check alone gives the same gain. Second, prefill-only decision models offer a competitive and comparatively inexpensive alternative to generative LLM judges, reducing cost with small drop in performance. Third, larger judges are not uniformly better: smaller models can be effective when their context capacity and decision thresholds are appropriate for the verification task. These results motivate future work of a modular verification approach in which deterministic checks handle structured facts and lightweight decision models assess conversational context.

\begin{table}[tp]
\centering
\small
\caption{The same \jevact{} write gate on $\tau^2$-Airline with different judges answering its compiled questions.
Each LLM judge uses its own label-free list of dropped questions. Judge cost per task is estimated from logged requests
and tokens; see Section~\ref{app:judge-cost}. ReAct is run separately for each block; only the first block's baseline is shown
(the GPT-4.1 baseline for the Kev-size block is 52.7\%). Best judge per block and column in \textbf{bold}.}
\label{tab:judge-choice}
\begin{adjustbox}{max width=\linewidth}
\begin{tabular}{@{}l l ccc ccc@{}}
\toprule
 & & \multicolumn{3}{c}{GPT-4.1} & \multicolumn{3}{c}{Haiku 4.5} \\
\cmidrule(l){3-5} \cmidrule(l){6-8}
Judge (\#param) & Type (model size) & \passhat{1} & \passhat{3} & \$ / task & \passhat{1} & \passhat{3} & \$ / task \\
\midrule
none (ReAct) & -- & 53.3 & 42.0 & -- & 55.3 & 36.0 & -- \\
Qwen3.8-27B (Kev-27B's base) & open LLM (27B) & 68.0 & 50.0 & 0.0027 & 64.7 & 48.0 & 0.0020 \\
Gemma-4-31B & open LLM (31B) & \textbf{72.0} & \textbf{56.0} & 0.0040 & 64.7 & 52.0 & 0.0037 \\
Claude Haiku 4.5 & API LLM & \textbf{72.0} & \textbf{56.0} & 0.0156 & 65.3 & 50.0 & 0.0127 \\
GPT-4.1 & API LLM & 67.3 & 52.0 & 0.0231 & \textbf{66.0} & \textbf{56.0} & 0.0195 \\
\midrule
\multicolumn{8}{@{}l}{\textit{Kev size}} \\
Kev-0.8B & decision model (0.8B) & 66.0 & 50.0 & 0.0003 & \textbf{68.7} & \textbf{58.0} & 0.0003 \\
Kev-9B & decision model (9B) & 63.3 & \textbf{56.0} & 0.0008 & 68.0 & \textbf{58.0} & 0.0006 \\
Kev-27B (\jevact{}) & decision model (27B) & \textbf{72.0} & \textbf{56.0} & 0.0017 & 68.0 & \textbf{58.0} & 0.0015 \\
\bottomrule
\end{tabular}

\end{adjustbox}
\end{table}

\begin{table}[tp]
\centering
\small
\caption{Judge size ablation. \jevact{} uses Kev-0.8B, Kev-9B or Kev-27B; compiled checks, thresholds and dropped
checklist steps (all set with Kev-27B) are fixed across arms. On Airline the judge answers the write gate's questions;
on Telecom it also answers the hand-off checklist. Best value of each row and metric is in \textbf{bold}.}
\label{tab:kevsize}
\begin{adjustbox}{max width=\linewidth}
\begin{tabular}{@{}ll cccc cccc@{}}
\toprule
 & & \multicolumn{4}{c}{\passhat{1}} & \multicolumn{4}{c}{\passhat{3}} \\
 & & & \multicolumn{3}{c}{\jevact{}, judge:} & & \multicolumn{3}{c}{\jevact{}, judge:} \\
\cmidrule(lr){4-6} \cmidrule(l){8-10}
Benchmark & Model & ReAct & Kev-0.8B & Kev-9B & Kev-27B & ReAct & Kev-0.8B & Kev-9B & Kev-27B \\
\midrule
\multirow{4}{*}{\begin{tabular}[c]{@{}l@{}}$\tau^2$ Airline \\\end{tabular}} & GPT-4.1 & 52.0 & 58.0 & 59.3 & \textbf{65.3} & 36.0 & 40.0 & 40.0 & \textbf{46.0} \\
 & Opus 4.6 & 79.3 & \textbf{86.0} & 83.3 & 84.0 & 72.0 & \textbf{82.0} & 76.0 & 76.0 \\
 & gpt-oss-120b & 56.7 & \textbf{62.0} & 58.7 & 58.0 & 42.0 & 44.0 & 46.0 & \textbf{50.0} \\
 & DeepSeek-V4-Pro & 77.3 & 79.3 & \textbf{80.7} & \textbf{80.7} & 68.0 & \textbf{70.0} & \textbf{70.0} & 66.0 \\
\midrule
\multirow{4}{*}{\begin{tabular}[c]{@{}l@{}}$\tau^2$ Telecom \\\end{tabular}} & GPT-4.1 & 25.7 & \textbf{54.4} & 52.3 & 49.7 & 11.4 & \textbf{27.2} & 23.7 & 17.5 \\
 & Opus 4.6 & 59.4 & 96.5 & \textbf{99.1} & 93.6 & 39.5 & 90.4 & \textbf{97.4} & 80.7 \\
 & gpt-oss-120b & 54.4 & \textbf{77.8} & \textbf{77.8} & 75.1 & 28.9 & 50.9 & \textbf{53.5} & 51.8 \\
 & DeepSeek-V4-Pro & 70.5 & 95.0 & \textbf{98.2} & 85.7 & 53.5 & 86.0 & \textbf{94.7} & 68.4 \\
\bottomrule
\end{tabular}
\end{adjustbox}
\end{table}

\section{Conclusion}
\label{sec:conclusion}

We introduced \jevact{}, a runtime gate that checks selected agent proposals before execution, combines executable checks over tool state with probability-based judgments of conversational context, and returns feedback for bounded revision. Compared with same-model ReAct baselines, the full harness shows its clearest gains on Telecom: all eight agents improve both at-least-once success and success across all three runs. On BFCL, all nine improve in three-run success, while at-least-once results are mixed. Airline results are smaller and mixed across agents. These findings support further study of decision-model gates, but do not establish uniform gains or isolate the judge's contribution.
Two caveats apply: with a gpt-oss-120b user simulator the Telecom gain does not appear (\passhat{3} changes by $-1.8$ points, 95\% CI $-7.9$ to $+4.4$), and on 260 labelled development write calls 42 of the 101 incorrect calls break no compiled question, because the questions encode the policy rather than the user's request.

\paragraph{Future work.}
The next step is to make the gate work across agent architectures rather than tying it to one ReAct loop. A common interface to proposed actions, tool state, and execution outcomes could support different action formats and orchestration frameworks. We also aim to automate more of the setup: deriving candidate checks from policies and tool schemas, generating and validating executable checks, and selecting thresholds and revision budgets on held-out traces. Such automation should account for both missed violations and unnecessary interventions, while retaining human escalation for uncertain or high-impact decisions. Broader evaluations should include more domains and frameworks, and report end-to-end latency, cost, and safety outcomes alongside task success and repeatability.

\bibliographystyle{plainnat}
\bibliography{references}

\begin{thebibliography}{29}
\providecommand{\natexlab}[1]{#1}
\providecommand{\url}[1]{\texttt{#1}}
\expandafter\ifx\csname urlstyle\endcsname\relax
  \providecommand{\doi}[1]{doi: #1}\else
  \providecommand{\doi}{doi: \begingroup \urlstyle{rm}\Url}\fi

\bibitem[{Anthropic}(2026)]{anthropicpricing}
{Anthropic}.
\newblock Pricing.
\newblock \url{https://platform.claude.com/docs/en/about-claude/pricing}, 2026.
\newblock Claude Haiku 4.5: \$1 / \$5 per million input / output tokens.
  Accessed 2026-10-14.

\bibitem[Barres et~al.(2025)Barres, Dong, Ray, Si, and
  Narasimhan]{barres2025tau}
Victor Barres, Honghua Dong, Soham Ray, Xujie Si, and Karthik Narasimhan.
\newblock {$\tau^2$}-bench: Evaluating conversational agents in a dual-control
  environment.
\newblock \emph{arXiv preprint arXiv:2506.07982}, 2025.

\bibitem[Chen et~al.(2021)Chen, Tworek, Jun, Yuan, Pinto, Kaplan, Edwards,
  Burda, Joseph, Brockman, et~al.]{chen2021evaluating}
Mark Chen, Jerry Tworek, Heewoo Jun, Qiming Yuan, Henrique Ponde De~Oliveira
  Pinto, Jared Kaplan, Harri Edwards, Yuri Burda, Nicholas Joseph, Greg
  Brockman, et~al.
\newblock Evaluating large language models trained on code.
\newblock \emph{arXiv preprint arXiv:2107.03374}, 2021.

\bibitem[Chen et~al.(2025)Chen, Kang, and Li]{chen2025shieldagent}
Zhaorun Chen, Mintong Kang, and Bo~Li.
\newblock Shieldagent: Shielding agents via verifiable safety policy reasoning.
\newblock \emph{arXiv preprint arXiv:2503.22738}, 2025.

\bibitem[Debenedetti et~al.(2026)Debenedetti, Shumailov, Fan, Hayes, Carlini,
  Fabian, Kern, Shi, Terzis, and Tram{\`e}r]{debenedetti2026defeating}
Edoardo Debenedetti, Ilia Shumailov, Tianqi Fan, Jamie Hayes, Nicholas Carlini,
  Daniel Fabian, Christoph Kern, Chongyang Shi, Andreas Terzis, and Florian
  Tram{\`e}r.
\newblock Defeating prompt injections by design.
\newblock In \emph{2026 IEEE Conference on Secure and Trustworthy Machine
  Learning (SaTML)}, pages 587--618. IEEE, 2026.

\bibitem[Deu{\ss}er et~al.(2026)Deu{\ss}er, Sparrenberg, and
  Sifa]{deusser2026evaluating}
Tobias Deu{\ss}er, Lorenz Sparrenberg, and Rafet Sifa.
\newblock Evaluating and benchmarking the system one model jev.
\newblock \emph{arXiv preprint arXiv:2609.37647}, 2026.

\bibitem[Kang et~al.(2026{\natexlab{a}})Kang, Yu, and
  Hwang]{kang2026policyguard}
Seongjae Kang, Taehyung Yu, and Sung~Ju Hwang.
\newblock Policyguard: A dialogue-grounded sub-agent verifier for policy
  adherence in llm agents.
\newblock \emph{arXiv preprint arXiv:2606.29225}, 2026{\natexlab{a}}.

\bibitem[Kang et~al.(2026{\natexlab{b}})Kang, Yu, and
  Hwang]{kang2026policyguide}
Seongjae Kang, Taehyung Yu, and Sung~Ju Hwang.
\newblock Policyguide: From guarding one action to guiding the whole workflow
  for policy-compliant llm agents.
\newblock \emph{arXiv preprint arXiv:2608.19861}, 2026{\natexlab{b}}.

\bibitem[Li et~al.(2026{\natexlab{a}})Li, He, Dai, and Xiao]{li2026toolfence}
Yanjie Li, Xiangyu He, Xuelong Dai, and Bin Xiao.
\newblock Toolfence: Fine-grained authorization for secure tool-using llm
  agents.
\newblock \emph{arXiv preprint arXiv:2609.37196}, 2026{\natexlab{a}}.

\bibitem[Li et~al.(2026{\natexlab{b}})Li, Miao, Krishnan, and
  Padman]{li2026jev}
Yubo Li, Yidi Miao, Ramayya Krishnan, and Rema Padman.
\newblock Jev-as-a-judge: Accept when confident, escalate when unsure.
\newblock \emph{arXiv preprint arXiv:2609.26550}, 2026{\natexlab{b}}.

\bibitem[Liu et~al.(2023)Liu, Iter, Xu, Wang, Xu, and Zhu]{liu2023g}
Yang Liu, Dan Iter, Yichong Xu, Shuohang Wang, Ruochen Xu, and Chenguang Zhu.
\newblock G-eval: Nlg evaluation using gpt-4 with better human alignment.
\newblock In \emph{Proceedings of the 2023 conference on empirical methods in
  natural language processing}, pages 2511--2522, 2023.

\bibitem[Madaan et~al.(2023)Madaan, Tandon, Gupta, Hallinan, Gao, Wiegreffe,
  Alon, Dziri, Prabhumoye, Yang, et~al.]{madaan2023self}
Aman Madaan, Niket Tandon, Prakhar Gupta, Skyler Hallinan, Luyu Gao, Sarah
  Wiegreffe, Uri Alon, Nouha Dziri, Shrimai Prabhumoye, Yiming Yang, et~al.
\newblock Self-refine: Iterative refinement with self-feedback.
\newblock \emph{Advances in neural information processing systems},
  36:\penalty0 46534--46594, 2023.

\bibitem[Mou et~al.(2026)Mou, Xue, Li, Liu, Zhang, Ye, and
  Shao]{mou2026toolsafe}
Yutao Mou, Zhangchi Xue, Lijun Li, Peiyang Liu, Shikun Zhang, Wei Ye, and Jing
  Shao.
\newblock Toolsafe: Enhancing tool invocation safety of llm-based agents via
  proactive step-level guardrail and feedback.
\newblock In \emph{Findings of the Association for Computational Linguistics:
  ACL 2026}, pages 37125--37153, 2026.

\bibitem[{OpenAI}(2026)]{openaipricing}
{OpenAI}.
\newblock {GPT-4.1} model pricing.
\newblock \url{https://developers.openai.com/api/docs/models/gpt-4.1}, 2026.
\newblock GPT-4.1: \$2 / \$8 per million input / output tokens, standard tier.
  Accessed 2026-10-14.

\bibitem[Palmer(2026)]{kev}
Jared Palmer.
\newblock {KEV}: Prefill-only decision models.
\newblock \url{https://github.com/jaredpalmer/kev}, 2026.
\newblock GitHub repository.

\bibitem[Patil et~al.(2025)Patil, Mao, Yan, Ji, Suresh, Stoica, and
  Gonzalez]{patil2025berkeley}
Shishir~G Patil, Huanzhi Mao, Fanjia Yan, Charlie Cheng-Jie Ji, Vishnu Suresh,
  Ion Stoica, and Joseph~E Gonzalez.
\newblock The berkeley function calling leaderboard (bfcl): From tool use to
  agentic evaluation of large language models.
\newblock In \emph{Forty-second International Conference on Machine Learning},
  2025.

\bibitem[Rao and Callison-Burch(2026)]{rao2026jev}
Delip Rao and Chris Callison-Burch.
\newblock Jev vs. llms as rubric judges: Cheaper, faster, and wrong in the same
  places.
\newblock \emph{arXiv preprint arXiv:2609.29769}, 2026.

\bibitem[Shinn et~al.(2023)Shinn, Cassano, Gopinath, Narasimhan, and
  Yao]{shinn2023reflexion}
Noah Shinn, Federico Cassano, Ashwin Gopinath, Karthik Narasimhan, and Shunyu
  Yao.
\newblock Reflexion: Language agents with verbal reinforcement learning.
\newblock \emph{Advances in neural information processing systems},
  36:\penalty0 8634--8652, 2023.

\bibitem[Shlomov et~al.(2026)Shlomov, Shoham, Oved, Levy, Marreed, Ship,
  Akrabi, Zeltyn, Yaeli, and Mashkif]{shlomov2026governance}
Segev Shlomov, Iftach Shoham, Alon Oved, Ido Levy, Sami Marreed, Harold Ship,
  Offer Akrabi, Sergey Zeltyn, Avi Yaeli, and Nir Mashkif.
\newblock Governance by construction for generalist agents.
\newblock In \emph{Proceedings of the ACM Conference on AI and Agentic
  Systems}, pages 1280--1287, 2026.

\bibitem[{TypeSafe AI}(2026)]{typesafe2026jev}
{TypeSafe AI}.
\newblock Introducing system one models and jev.
\newblock \url{https://typesafe.ai/blog/introducing-system-one-models-and-jev},
  2026.
\newblock Released September 15, 2026.

\bibitem[Wang et~al.(2026)Wang, Poskitt, and Sun]{wang2026agentspec}
Haoyu Wang, Christopher~M Poskitt, and Jun Sun.
\newblock Agentspec: Customizable runtime enforcement for safe and reliable llm
  agents.
\newblock In \emph{Proceedings of the 2026 IEEE/ACM 48th International
  Conference on Software Engineering}, pages 2938--2950, 2026.

\bibitem[Wang et~al.(2024)Wang, Li, Chen, Cai, Zhu, Lin, Cao, Kong, Liu, Liu,
  et~al.]{wang2024large}
Peiyi Wang, Lei Li, Liang Chen, Zefan Cai, Dawei Zhu, Binghuai Lin, Yunbo Cao,
  Lingpeng Kong, Qi~Liu, Tianyu Liu, et~al.
\newblock Large language models are not fair evaluators.
\newblock In \emph{Proceedings of the 62nd annual meeting of the association
  for computational linguistics (volume 1: Long papers)}, pages 9440--9450,
  2024.

\bibitem[Xiang et~al.(2024)Xiang, Zheng, Li, Hong, Li, Xie, Zhang, Xiong, Xie,
  Yang, et~al.]{xiang2024guardagent}
Zhen Xiang, Linzhi Zheng, Yanjie Li, Junyuan Hong, Qinbin Li, Han Xie, Jiawei
  Zhang, Zidi Xiong, Chulin Xie, Carl Yang, et~al.
\newblock Guardagent: Safeguard llm agents by a guard agent via
  knowledge-enabled reasoning.
\newblock \emph{arXiv preprint arXiv:2406.09187}, 2024.

\bibitem[Yabusaki(2026)]{yabusaki2026verifier}
Hitoshi Yabusaki.
\newblock Verifier-grounded runtime mediation for llm agent procedural
  compliance.
\newblock In \emph{2026 IEEE International Conference on Software Services
  Engineering (SSE)}, pages 34--44. IEEE, 2026.

\bibitem[Yao et~al.(2022)Yao, Zhao, Yu, Du, Shafran, Narasimhan, and
  Cao]{yao2022react}
Shunyu Yao, Jeffrey Zhao, Dian Yu, Nan Du, Izhak Shafran, Karthik Narasimhan,
  and Yuan Cao.
\newblock React: Synergizing reasoning and acting in language models.
\newblock \emph{arXiv preprint arXiv:2210.03629}, 2022.

\bibitem[Yao et~al.(2024)Yao, Shinn, Razavi, and Narasimhan]{yao2024tau}
Shunyu Yao, Noah Shinn, Pedram Razavi, and Karthik Narasimhan.
\newblock {$\tau$}-bench: A benchmark for tool-agent-user interaction in
  real-world domains.
\newblock \emph{arXiv preprint arXiv:2406.12045}, 2024.

\bibitem[Zheng et~al.(2023)Zheng, Chiang, Sheng, Zhuang, Wu, Zhuang, Lin, Li,
  Li, Xing, et~al.]{zheng2023judging}
Lianmin Zheng, Wei-Lin Chiang, Ying Sheng, Siyuan Zhuang, Zhanghao Wu, Yonghao
  Zhuang, Zi~Lin, Zhuohan Li, Dacheng Li, Eric Xing, et~al.
\newblock Judging llm-as-a-judge with mt-bench and chatbot arena.
\newblock \emph{Advances in neural information processing systems},
  36:\penalty0 46595--46623, 2023.

\bibitem[{ZJUNLP}(2026)]{jevloop2026}
{ZJUNLP}.
\newblock {JevLoop}: {Acta}: A decision-centric runtime for agents.
\newblock \url{https://github.com/zjunlp/JevLoop}, 2026.
\newblock GitHub repository, accessed October 14, 2026.

\bibitem[Zwerdling et~al.(2025)Zwerdling, Boaz, Rabinovich, Uziel, Amid, and
  Tavor]{zwerdling2025towards}
Naama Zwerdling, David Boaz, Ella Rabinovich, Guy Uziel, David Amid, and
  Ateret~Anaby Tavor.
\newblock Towards enforcing company policy adherence in agentic workflows.
\newblock In \emph{Proceedings of the 2025 Conference on Empirical Methods in
  Natural Language Processing: Industry Track}, pages 595--606, 2025.

\end{thebibliography}

\appendix

\section{Appendix}
\label{sec:appendix}

\subsection{Offline compilation of the questions and checks}
\label{app:compile}

This section expands Section~\ref{sec:compile}: how the compiler turns a policy into questions, rules, code, and a hand-off
checklist. It includes the available prompt templates and representative outputs.

\paragraph{Inputs and outputs.}
Claude Opus~4.6 makes a separate prompted call for each artifact-generation step; these are single-turn calls, though
the client may retry transport failures. Steps 1, 3, 4, and the checklist call in step 6 allow up to 16,384 output
tokens; the rule call allows 8,192. Telecom checklist generation also makes a separate prerequisite-extraction call
with a 2,048-token limit. Step 5 is a local filter over Kev's answers, not a Claude call. Inputs include the domain
policy as the agent receives it, tool schemas, and, for code generation, one logged call and result per read tool to
show the JSON shapes. \Cref{tab:compile-steps} lists the steps. Steps 1--5 build the write gate; step 6 builds the
hand-off checklist.

\begin{table}[h]
\centering
\small
\caption{The offline compilation workflow. ``Logged'' calls and hand-offs come from development runs. Telecom step 6
also includes a separate prerequisite-extraction call.}
\label{tab:compile-steps}
\begin{tabular}{@{}p{0.2\textwidth}p{0.29\textwidth}p{0.29\textwidth}p{0.1\textwidth}@{}}
\toprule
Step & Reads & Writes & Labels \\
\midrule
1. Questions & policy; write-tool schemas & questions per write tool & none \\
2. Rule & policy; step-1 questions & allow rule per write tool & none \\
3. Code & policy, schemas; one read result per tool & \texttt{derive\_<tool>} function per write tool & none \\
4. Repair & code and flags on logged calls & repaired code & none \\
5. Filter & Kev answers on logged write calls & questions to drop & none \\
6a. Prerequisites & policy; tool names (Telecom only) & prerequisites & none \\
6b. Hand-off check & policy, prerequisites; Kev's logged answers & problems, steps; drop steps & outcomes \\
\bottomrule
\end{tabular}
\end{table}

In the prompts below, text in angle brackets (\texttt{<\ldots>}) is filled by code; everything else is sent verbatim,
except that line breaks inside a paragraph are removed here for typesetting.

\paragraph{Step 1: questions per write tool.}
The prompt describes the reader the questions are for: a small model that sees the conversation, cannot compute, and
answers yes/no with a probability. It asks for questions that each test one fact and together cover the policy's
conditions on the call, the provenance of every argument, whether the user wants the action, and the required steps
before it. Two output fields matter at run time. \texttt{bad\_if} gives the answer that means a violation, so the gate
knows each question's polarity. Arguments are written as placeholders, \texttt{\{reservation\_id\}}, which code fills
with the proposed call's values before Kev reads the question. The prompt also asks the compiler not to make the
reviewer compute anything; arithmetic is left to step 3.
\begin{Verbatim}
You are compiling a customer-service agent's policy into atomic checks that an independent reviewer will answer before each database-changing tool call runs.

The reviewer is a small reading model. It sees the whole conversation: the user's messages, the agent's messages, every tool call and its JSON result, and the policy. Dates in the conversation are annotated with their offset from the current time, and lists of options in tool results come with pre-computed rankings (lowest to highest price, etc.). It cannot do arithmetic itself and cannot see anything outside the conversation. It answers yes/no questions with a calibrated probability.

For EACH write tool below, write 12 to 20 atomic yes/no questions such that, together, they cover every policy rule and every way the call could be wrong for this user:
- conditions the policy puts on the action (eligibility, required preconditions, what must be checked first);
- whether each argument value is right: grounded in a tool result, what the user actually asked for or explicitly chose (not just agreed to after the agent proposed it), and consistent with the policy's formulas;
- whether the action is wanted at all (the user asked for it; not something the user declined or did not request);
- required steps before acting (records retrieved, explicit confirmation of the details).
Each question must test ONE fact, be answerable from the conversation, and refer to the proposed call's arguments as {arg_name} (e.g. {reservation_id}). Prefer facts read from tool results over the user's claims. Do not ask the reviewer to compute anything.

Return only JSON: {"<tool name>": [{"id": "<short_snake_case>", "question": "...", "bad_if": "yes" or "no", "rule": "<the policy rule it checks, short>"}, ...], ...}

POLICY:
<the domain policy, verbatim>

WRITE TOOLS (OpenAI schemas):
<the write tools' OpenAI function schemas, as JSON>
\end{Verbatim}
For airline this produced 86 questions over six write tools (12 to 21 per tool); for telecom, 75. Part of the
compiled list for \texttt{cancel\_reservation}:
\begin{Verbatim}
{"id": "reservation_id_correct", "bad_if": "no",
 "question": "Does {reservation_id} match a reservation ID associated with the user as confirmed by tool results?"}
{"id": "no_flown_segments", "bad_if": "no",
 "question": "Based on tool results, have zero flight segments in {reservation_id} already been flown (status 'flying' or past departure)?"}
{"id": "cancellation_eligible_24hr", "bad_if": "no",
 "question": "Was the booking for {reservation_id} made within the last 24 hours according to tool results?"}
{"id": "at_least_one_cancel_condition_met", "bad_if": "no",
 "question": "Is at least one cancellation eligibility condition met: booked within 24hrs, airline cancelled, business cabin, or insurance-covered reason?"}
{"id": "user_requested_cancellation", "bad_if": "no",
 "question": "Did the user actually request to cancel this reservation (not just inquire about it)?"}
{"id": "user_confirmed_cancellation", "bad_if": "no",
 "question": "Did the agent list the cancellation details and did the user explicitly confirm (say yes) before this tool call?"}
\end{Verbatim}

\paragraph{Step 2: the decision rule.}
A second call reads the policy and the step-1 questions (ids and text only) and writes, per tool, one Python boolean
expression over the question ids that is true when the call is allowed. This is where alternatives are combined: the
policy allows a cancellation if any one of four conditions holds, which a list of independent questions cannot express.
\begin{Verbatim}
Below is a customer-service policy and, for each database-changing tool, a list of yes/no questions an independent reviewer answers about a proposed call (each with an id).

For each tool, write the policy's decision as ONE Python boolean expression over the question ids, True when the call is allowed and correct. Each id evaluates to True if the reviewer answered yes. Use `and`, `or`, `not` and parentheses. Combine alternative eligibility conditions with `or` (e.g. any one of several cancellation conditions is enough). Leave out questions that are not needed for the decision or that ask about things the policy does not require. Return only JSON: {"<tool name>": "<expression>", ...}

POLICY:
<the domain policy, verbatim>

QUESTIONS:
<the tool's questions from the first compile, as JSON {tool: [{id, question}, ...]}>
\end{Verbatim}
The compiled rule for \texttt{cancel\_reservation} keeps the summary eligibility question and leaves out the four
single-condition ones:
\begin{Verbatim}
user_id_obtained and reservation_id_correct and reservation_retrieved and no_flown_segments and
cancellation_reason_obtained and at_least_one_cancel_condition_met and user_requested_cancellation and
user_confirmed_cancellation
\end{Verbatim}
At run time each id is true unless Kev answers a confident \emph{no} ($P(\text{yes}) \le 0.05$), and the rule is
evaluated with Python's \texttt{eval} and no builtins. If it is false, the feedback lists the questions that failed.

\paragraph{Step 3: code for what the records decide.}
Questions that need arithmetic or date comparisons (free baggage allowance, hours since booking, certificate amounts)
are what Kev answers worst. A third call writes one Python function per write tool that reads the proposed arguments
and every earlier tool result, computes the facts the policy depends on, and returns the violations the records make
certain. The prompt restricts the code to the standard library; it is executed with a whitelist of builtins and
imports, and an exception counts as no violation.
\begin{Verbatim}
You are writing helper code for an independent reviewer that checks a customer-service agent's database-changing tool calls against the policy below. The reviewer is a reading model: it cannot do arithmetic, compare dates, look up tables or compute minima. Your code does that for it.

For EACH write tool listed, write a Python function

    def derive_<tool name>(args: dict, results: list) -> dict:

- `args`: the proposed call's arguments.
- `results`: every earlier tool call in the conversation, oldest first, as dicts {"tool": <tool name>, "args": <its arguments>, "result": <its parsed JSON result, or the raw string>}. Example results of each read tool are given below, so you know the JSON shapes.
- Return a dict of COMPUTED FACTS the policy's rules for this tool depend on, as short readable strings or numbers, e.g. allowances and totals the policy defines, the value an argument should have by the policy's formula next to the value the call uses, time elapsed since a timestamp vs the policy's current time, balances vs amounts, counts, which earlier records the call's ids refer to and their relevant fields. Name each fact clearly (e.g. "expected_nonfree_baggages", "hours_since_booking").
- Also return the key "violations": a list of short strings, one per policy rule this call clearly breaks according to the RECORDS (tool results) and your computations, e.g. "nonfree_baggages should be 0 (gold, economy, 1 passenger: 3 free; total 2), call has 1", or "reservation created 312 h before now; not within 24 h, not business, no insurance, no airline cancellation: not eligible for cancellation". Only include a violation when the records make it certain; if a needed record is missing, or the rule depends on what the user said or wants, do not report a violation (a reviewer handles those). An empty list means no violation found.
- Use only the standard library modules json, math, datetime (already imported; no other imports). Be defensive: if something is missing, skip that fact. Never raise.

Return only the Python code, no explanations.

POLICY:
<the domain policy, verbatim>

WRITE TOOLS:
<the write tools' OpenAI function schemas, as JSON>

EXAMPLE RESULTS OF READ TOOLS (from logs, truncated):
<one logged call and result per read tool, as JSON {tool: {args, result}}; each result cut to 1,500 characters>
\end{Verbatim}
An excerpt of the compiled airline code for \texttt{update\_reservation\_baggages} (helper functions omitted):
\begin{Verbatim}
free_per_pax = _free_bags_per_passenger(membership, cabin)
if free_per_pax is not None:
    total_free = free_per_pax * num_passengers
    expected_nonfree = max(0, new_total - total_free)
    if new_nonfree != expected_nonfree:
        violations.append(
            f"nonfree_baggages should be {expected_nonfree} "
            f"({membership}, {cabin}, {num_passengers} passenger(s): "
            f"{total_free} free; new total {new_total}), call has {new_nonfree}")
\end{Verbatim}

\paragraph{Step 4: one repair round.}
The step-3 code is run on the write calls in the logs of train-split tasks. Up to 16 calls on which it reported a
violation, at most one per tool and first violation, are shown back to the compiler with the records the code read.
No gold action or outcome is shown; the compiler judges each reported violation against the policy and the records
and fixes the bugs it finds (for example, reading an error string such as \texttt{"Error: User not found"} as a user
record, or using a reservation that a later tool result had superseded). On held-out airline tasks this round cut
the code's false alarms from 11.3\% to 1.4\% of correct write calls.
\begin{Verbatim}
Below is Python code you wrote earlier that computes, for a customer-service agent's proposed database-changing tool call, the facts the policy depends on and a list of "violations" that the records make certain.

It was run on calls from real conversations. Here are calls where it reported violations, each with the call, the violations it reported, and the earlier tool calls and results it read (oldest first, truncated). For each one, check against the policy and the records whether the reported violation is real. Where it is not (a bug: e.g. it read an error message as a record, used a stale record that a later tool result superseded, misread a field, applied the wrong rule), fix the code. Keep every check that is right. Return the complete corrected code only.

POLICY:
<the domain policy, verbatim>

CODE:
```python
<the compiled module from the previous step>
```

FLAGGED CALLS:
<up to 16 logged calls (train-split tasks) on which the code reported a violation, at most one per tool and first violation; each given as the call, the violations it reported, and the last 8 earlier tool calls with their results (each result cut to 700 characters)>
\end{Verbatim}

\paragraph{Step 5: label-free question filter.}
Kev-27B answers every compiled question on every write call in the development logs, with the same state it sees
online (the policy, the conversation, every tool result with dates annotated and lists ranked, and the proposed call).
A question \emph{fires} on a call when its answer is a confident violation. Any question that fires on more than 10\%
of its tool's logged calls is dropped: real violations are rare, so a question that fires that often is asking
something Kev cannot read from the conversation or the policy does not require. A dropped question is never reported
and counts as \emph{yes} in the rule. The thresholds ($P \le 0.05$, 10\%) were set once on airline and reused unchanged.
This drops 17 of 86 Airline questions and 5 of 75 Telecom questions (175 logged Telecom write calls).

\paragraph{Step 6: the hand-off checklist.}
The hand-off prompt asks for the problem types the policy troubleshoots and, for each, the steps the agent must try
before handing off. Each step carries a \emph{settled} question, the problems that require it (prerequisites
included) and the agent tools that perform it. The prompt below describes the checklist-generation call.
\begin{Verbatim}
You are compiling a customer-service agent's policy into a checklist for an independent reviewer.

The reviewer is called at the moment the agent wants to END its attempt to solve the user's problem: hand the user off to a human agent, or tell the user nothing more can be done. The policy asks the agent to try every relevant resolution step first. The reviewer reads the conversation as the agent saw it (the user's messages, the agent's messages, the agent's own tool calls and their results) and flags a REQUIRED step that is still OPEN.

1. "problems": the kinds of problem the policy has troubleshooting for, as {"<id>": "<one-line description>"}.
2. "steps": every resolution / troubleshooting step the agent must go through (itself with a tool, or by guiding the user on their device) before handing off. Include steps that start by establishing a fact (e.g. asking the user something) when the policy's later steps depend on it: the agent cannot skip them because the user did not volunteer the fact. For each step:
   - "id": short snake_case id;
   - "step": what the agent must do, one sentence, naming the tool or device action (and the question to ask, if any);
   - "problems": the problem ids it is required for, INCLUDING problems for which the policy makes it a prerequisite (e.g. a step for problem A is also required for problem B if B's troubleshooting says to fix A first);
   - "settled": a yes/no question, YES when the conversation shows the step is settled: it was tried, OR the setting or condition it checks was verified to be fine, OR the agent established that the step does not apply (e.g. the user answered a question that rules it out). NO when the conversation never addressed it. A condition that was never discussed is NOT ruled out: if nobody asked (e.g. whether the user is traveling), the step is open. Do not put an "or <condition> does not hold" escape in the question unless it says "the conversation established that";
   - "agent_tools": the agent tools (names from the list below) that perform or verify the step, [] if none.

Cover the policy's own steps, in its order; skip identity verification and pure information requests. Reply with one JSON object: {"problems": {...}, "steps": [...]}.

AGENT TOOLS: <the names of the agent's tools, comma-separated>

POLICY:
<the domain policy, verbatim>
\end{Verbatim}

Online, Kev answers in one request which problem the user has (one of the compiled types or \emph{other}),
whether the user confirmed the problem is fixed, and every step's settled question. A step is open if the
probability of the problems that require it is at least 0.5 and $P(\text{settled}) \le 0.5$.

To select checklist steps for dropping, Kev-27B answered the checklist questions at the first hand-off in each episode across six Telecom
development runs (GPT-4.1 and Opus~4.6). We used the 74 train-split tasks (217 hand-offs, 36 in solved episodes): any
step flagged on more than 5\% of solved hand-offs was dropped, since a correct hand-off should pass. This removed seven
steps: airplane mode, APN settings, SIM status, line suspension, overdue bills, contract end date, and resuming the line
after payment. Both thresholds remained at 0.5. The 40 test-split tasks did not determine these seven drops. They were
held out only for the filter; the checklist itself was revised after a results review (Step 6).

\paragraph{BFCL.}
Nothing is compiled for BFCL multi-turn: it has no written policy, and its two checks ask the same fixed questions
on every task (Section~\ref{sec:checks}).
\subsection{Judge size on identical hand-off states (offline)}
\label{app:offline-judges}

We replayed the 609 telecom hand-off states from the Kev-27B runs through each judge, holding the checklist,
thresholds, and dropped steps fixed. This isolates judge differences from changes in the agent, user simulator, or
episode. A catch is a flag naming a required fix that remains undone; a false alarm names a fix the task does not
need.

\begin{table}[h]
\centering
\small
\caption{Offline replay of the same 609 telecom hand-off states for each Kev size.}
\label{tab:offline-judges}
\begin{tabular}{@{}lccc@{}}
\toprule
 & Kev-27B & Kev-9B & Kev-0.8B \\
\midrule
Hand-offs flagged & 371 & 369 & 342 \\
Same flag/pass decision as Kev-27B & -- & 97.0\% & 90.6\% \\
Required fixes caught (of 533) & 319 & 316 & 299 \\
Unneeded fixes flagged (of 245) & 193 & 195 & 170 \\
Failed episodes incorrectly marked fixed (of 380) & 0 & 0 & 11 \\
\bottomrule
\end{tabular}
\end{table}

Kev-9B closely matches Kev-27B in this replay. Kev-0.8B flags less, catching fewer required fixes but also raising
fewer false alarms. Neither smaller judge shows an overall decision-quality advantage that would explain its higher
online Telecom scores in Table~\ref{tab:kevsize}.

\subsection{Judge input length}
\label{app:judge-inputs}

Kev-0.8B and Kev-9B are validated up to 8,192 tokens; Kev-27B is validated up to 65,536~\citep{kev}. In the Kev-27B
runs, 41\% of Airline write-check inputs exceed 8,192 tokens, compared with 4\% of Telecom hand-off inputs. A write
check includes the policy and tool results as well as the conversation, while a hand-off check reads the conversation
and checklist.

\subsection{Judge size on labeled write calls (offline)}
\label{app:write-judges}

To separate ability from calibration, we replayed the same 260 labeled Airline write calls (101 bad, 159 good) through
each judge using the deployed questions and dropped-question list. Calls were split at 8,192 input tokens (180 short,
80 long). The table reports ranking quality without a threshold, catches and false alarms at the deployed threshold,
and results at each judge's best threshold subject to Kev-27B's false-alarm count. These best thresholds use the
labels and are therefore an upper bound for the smaller judges.

\begin{table}[h]
\centering
\small
\caption{Offline comparison on the same 260 labeled Airline write calls.}
\label{tab:write-judges}
\begin{adjustbox}{max width=\linewidth}
\begin{tabular}{@{}l ccc ccc@{}}
\toprule
 & \multicolumn{3}{c}{input $\le$ 8,192: 69 bad, 111 good} & \multicolumn{3}{c}{input $>$ 8,192: 32 bad, 48 good} \\
\cmidrule(lr){2-4} \cmidrule(l){5-7}
Kev size & 27B & 9B & 0.8B & 27B & 9B & 0.8B \\
\midrule
\multicolumn{7}{@{}l}{\textit{A. Ranking, no threshold (ability)}} \\
AUROC & 0.75 & 0.70 & 0.63 & 0.85 & 0.61 & 0.50 \\
\quad vs.\ 27B, 95\% CI & -- & $-$0.15 to +0.04 & $-$0.23 to $-$0.02 & -- & $-$0.37 to $-$0.12 & $-$0.50 to $-$0.19 \\
\midrule
\multicolumn{7}{@{}l}{\textit{B. Deployed threshold, $P(\text{yes}) \le 0.05$, set with Kev-27B}} \\
Caught/false alarms & 21/0 & 8/0 & 0/0 & 11/2 & 2/1 & 0/0 \\
\midrule
\multicolumn{7}{@{}l}{\textit{C. Own best threshold, false alarms at most Kev-27B's}} \\
Threshold & 0.06 & 0.10 & 0.15 & 0.06 & 0.10 & 0.15 \\
Caught/false alarms & 21/0 & 16/0 & 8/1 & 12/2 & 2/2 & 1/1 \\
\bottomrule
\end{tabular}

\end{adjustbox}
\end{table}

On long inputs, Kev-9B catches 2 of 32 bad calls at either threshold, versus 11 for Kev-27B; its AUROC is 0.61 versus
0.85. On short inputs, recalibrating Kev-9B from 0.05 to 0.10 raises catches from 8 to 16 (Kev-27B catches 21),
without adding a false alarm. Kev-0.8B ranks worse even on short inputs (AUROC 0.63 versus 0.75). These results
suggest that the size gap reflects both input length and calibration, while the smallest judge also has lower ranking
quality.

\subsection{Token usage}
\label{app:tokens}

Table~\ref{tab:tokens} reports mean tokens per run for the main model and Kev in the runs of Table~\ref{tab:main}. On
$\tau^2$-bench, usage comes from run logs; agent drafts sent back for revision are not recorded, so their count is
reported separately. On BFCL, Kev requests were rebuilt from logged conversations and counted with the model
tokenizer. Requests that could not be rebuilt are omitted and counted in the table.

\begin{table}[h]
\centering
\small
\caption{Mean tokens per run, in thousands, for the runs in Table~\ref{tab:main}. Agent usage is logged; Kev input is logged
on $\tau^2$-bench and rebuilt on BFCL. Drafts sent back and checks not rebuilt are counts per run.}
\label{tab:tokens}
\begin{tabular}{@{}ll rr rrrrr@{}}
\toprule
 & & \multicolumn{2}{c}{ReAct} & \multicolumn{5}{c}{\jevact{}} \\
\cmidrule(lr){3-4} \cmidrule(l){5-9}
 & & \multicolumn{2}{c}{Model} & \multicolumn{2}{c}{Model} & Kev & Drafts & Not \\
Benchmark & Model & in & out & in & out & in & sent back & rebuilt \\
\midrule
\multirow{6}{*}{\begin{tabular}[c]{@{}l@{}}$\tau^2$-bench \\ airline \\ {\scriptsize 50 tasks $\times$ 3}\end{tabular}} & GPT-4.1 & 58.4 & 0.9 & 60.1 & 0.9 & 9.0 & 0.39 & 0 \\
 & Opus 4.6 & 64.6 & 1.5 & 64.1 & 1.5 & 6.7 & 0.08 & 0 \\
 & gpt-oss-20b & 50.4 & 3.0 & 56.8 & 3.4 & 5.2 & 0.19 & 0 \\
 & gpt-oss-120b & 44.1 & 2.5 & 43.8 & 2.6 & 6.5 & 0.36 & 0 \\
 & Qwen3.8-27B & 79.8 & 8.2 & 80.3 & 8.5 & 6.4 & 0.03 & 0 \\
 & DeepSeek-V4-Pro & 64.0 & 1.8 & 63.2 & 1.8 & 9.5 & 0.11 & 0 \\
\midrule
\multirow{8}{*}{\begin{tabular}[c]{@{}l@{}}$\tau^2$-bench \\ telecom \\ {\scriptsize 114 tasks $\times$ 3}\end{tabular}} & GPT-4.1 & 130.6 & 1.2 & 167.3 & 1.4 & 16.8 & 1.44 & 0 \\
 & Haiku 4.5 & 138.8 & 2.1 & 184.9 & 2.7 & 21.0 & 1.91 & 0 \\
 & Opus 4.6 & 171.7 & 1.7 & 201.4 & 1.8 & 12.7 & 0.61 & 0 \\
 & Opus 5.5 & 178.1 & 2.2 & 183.0 & 2.2 & 9.6 & 0.10 & 0 \\
 & gpt-oss-20b & 157.3 & 9.1 & 167.9 & 9.4 & 22.1 & 1.30 & 0 \\
 & gpt-oss-120b & 117.7 & 4.2 & 139.0 & 5.1 & 24.1 & 1.38 & 0 \\
 & Qwen3.8-27B & 143.6 & 8.0 & 143.6 & 8.0 & 12.7 & 0.23 & 0 \\
 & DeepSeek-V4-Pro & 199.6 & 1.8 & 207.3 & 1.8 & 15.8 & 0.63 & 0 \\
\midrule
\multirow{9}{*}{\begin{tabular}[c]{@{}l@{}}BFCL \\ multi-turn \\ {\scriptsize 800 entries $\times$ 3}\end{tabular}} & GPT-4.1 & 44.1 & 0.5 & 46.5 & 0.5 & 3.2 & 0.02 & 0.02 \\
 & Opus 4.6 & 74.0 & 1.4 & 74.9 & 1.4 & 4.4 & 0.11 & 0.19 \\
 & gpt-oss-20b & 49.7 & 2.2 & 49.9 & 2.2 & 3.5 & 0.03 & 0.04 \\
 & gpt-oss-120b & 49.5 & 1.6 & 50.6 & 1.7 & 3.7 & 0.04 & 0.08 \\
 & Qwen3.8-27B & 88.8 & 3.3 & 87.8 & 3.3 & 5.5 & 0.11 & 0.19 \\
 & DeepSeek-V4-Pro & 81.9 & 1.4 & 81.9 & 1.4 & 5.9 & 0.13 & 0.22 \\
 & Llama 3.3 70B & 111.3 & 0.6 & 116.5 & 0.7 & 5.2 & 0.05 & 0.09 \\
 & Mistral Small 3.2 24B & 51.7 & 0.7 & 54.5 & 0.7 & 3.0 & 0.02 & 0.03 \\
 & Gemma 4 31B & 62.4 & 0.5 & 62.3 & 0.5 & 3.6 & 0.05 & 0.09 \\
\bottomrule
\end{tabular}

\end{table}

\subsection{Judge cost}
\label{app:judge-cost}

The judge costs in Table~\ref{tab:judge-choice} are estimates per episode, based on logged requests and tokens. API costs
use public standard-tier prices as of October 14, 2026 (GPT-4.1: \$2/\$8 and Haiku~4.5: \$1/\$5 per million
input/output tokens)~\citep{openaipricing,anthropicpricing}; these are not invoiced costs. Missing API output-token
counts are imputed from offline mean outputs.

For self-hosted judges, GPU time per request was measured offline on an A100-80GB (bf16) with four concurrent requests:
0.53 seconds for Kev-0.8B, 1.22 for Kev-9B, 2.72 for Kev-27B, 4.02 for Qwen3.8-27B, and 6.10 for Gemma-4-31B.
Costs use these measurements with an assumed \$2 per GPU-hour and full utilization. Estimates exclude the agent, user
simulator, and one-time compilation. GPU cost comparisons depend on serving stack and utilization; open LLMs ran
without a thinking phase.

\end{document}